\documentclass[runningheads]{llncs}
\usepackage{enumerate}
\usepackage{graphicx}
\usepackage{amsmath}
\usepackage{amsfonts}
\begin{document}
\title{Optimal Rejection Function Meets Character Recognition Tasks}

\author{Xiaotong Ji\inst{1}\and
Yuchen Zheng\inst{1}\orcidID{0000-0003-3093-6929} \and\\ 
Daiki Suehiro\inst{1,2}\orcidID{0000-0001-8901-9063} \and
Seiichi Uchida\inst{1}\orcidID{0000-0001-8592-7566}}
\authorrunning{Ji. et al.}
%
\institute{Kyushu University, 744 Motooka Nishi-ku, Fukuoka 819-0395, Japan\\
\email{\{xiaotong.ji, yuchen\}@human.ait.kyushu-u.ac.jp}\\
\email{\{suehiro, uchida\}@ait.kyushu-u.ac.jp}\and
RIKEN, 2-1 Hirosawa, Wako, Saitama 351-0198, Japan}
\maketitle              
\begin{abstract}
In this paper, we propose an optimal rejection method for rejecting ambiguous samples by a rejection function. This rejection function is trained together with a classification function under the framework of Learning-with-Rejection (LwR). The highlights of LwR are: (1) the rejection strategy is not heuristic but has a strong background from a machine learning theory, and (2) the rejection function can be trained on an arbitrary feature space which is different from the feature space for classification. The latter suggests we can choose a feature space which is more suitable for rejection. Although the past research on LwR focused only its theoretical aspect, we propose to utilize LwR for practical pattern classification tasks. Moreover, we propose to use features from different CNN layers for classification and rejection. Our extensive experiments of notMNIST classification and character/non-character classification demonstrate that the proposed method achieves better performance than traditional rejection strategies.

\end{abstract}
\section{Introduction}
According to general performance improvement in various recognition tasks  \cite{schroff2015facenet,sun2015deepid3,rautaray2015vision}, many users may expect that most test samples should be classified correctly. In fact, recent classifiers use forced decision-making strategies, where any test sample is always classified into one of the target classes  \cite{rouhi2015benign,yang2016hierarchical}. However, even with the recent classifiers, it is theoretically impossible to classify test samples correctly in several cases. For example, we cannot guarantee the correct recognition of a sample from the overlapping region of class distributions. 
\par
\begin{figure}[t!]
\begin{center}
\includegraphics[width=\columnwidth]{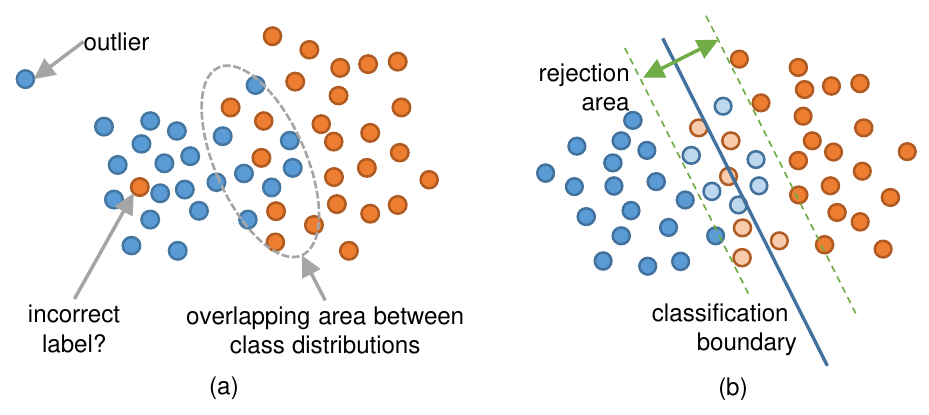}\\[-2mm]
\caption{(a)~Three main targets for rejection. (b)~Typical heuristics for rejecting the samples in the overlapping area of class distributions. This method assumes that we already have an accurate classification boundary --- but, how can we reject ambiguous samples when we train the classification boundary?}
\label{fig:reject-targets}
\end{center}
\end{figure}

As a practical remedy to deal with those ambiguous samples, recognition with a rejection option has been used so far \cite{condessa2016supervised,marinho2017novel}. As shown in Figure~\ref{fig:reject-targets}~(a), there are three main targets for rejection: outliers, samples with incorrect labels, and samples in an overlapping area of class distributions. Among them, outliers and incorrect labels are detected by anomaly detection methods. The overlapping area is a more common reason to introduce the rejection option into the training stage as well as the testing stage, since the samples in the overlapping area are inherently distinguishable and thus better to be treated as ``don't-care'' samples.\par 

In the long history of pattern recognition, researchers have tried to reject ambiguous samples in various ways. The most typical method is to reject samples around the classification boundary. Figure~\ref{fig:reject-targets}~(b) shows a naive rejection method for a support vector machine (SVM) for rejecting samples in the overlapping area. If the distance between a sample and the classification boundary is smaller than a threshold, the sample is going to be rejected~\cite{bartlett2008classification}.
\par
\setlength{\belowcaptionskip}{-0.5cm}

Although this simple rejection method is often employed for practical applications, it has a clear problem that it needs an accurate classification boundary in advance to rejection. In other words, it can reject only test samples. Ambiguous samples in the overlapping area also exist in the training set and they will badly affect the classification boundary. This fact implies that we need to introduce a more sophisticated rejection method where the rejection criterion should be ``co-optimized'' with the classification boundary in the training stage.
\par


\begin{figure}[t!]
\begin{center}
\includegraphics[width=110mm]{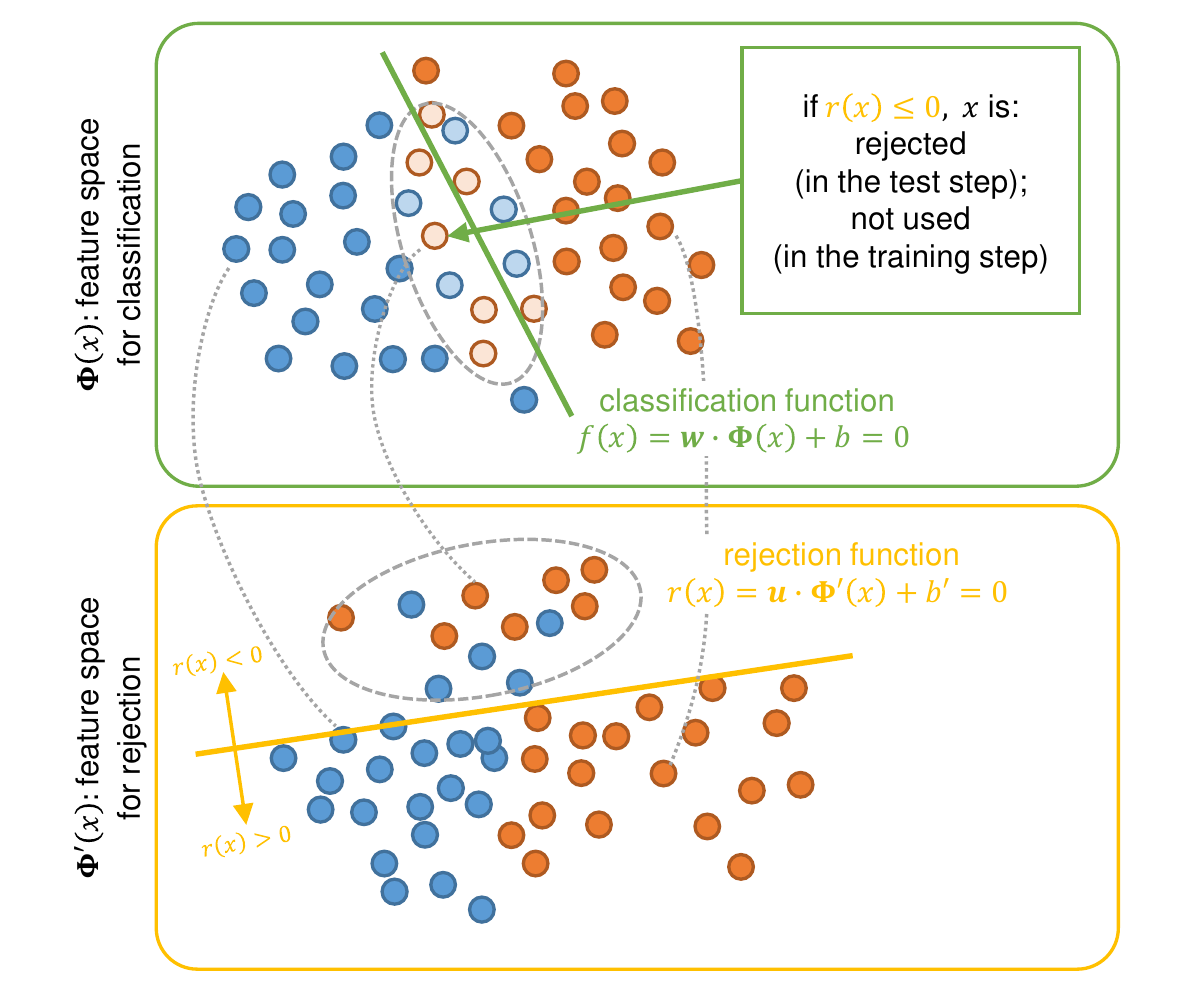}\\[-2mm]
\caption{An overview of {\em learning-with-rejection} for a two-class problem. The rejection function $r(x)$ is co-trained with the classification function $f(x)$. Note that different feature spaces are used for optimizing $r(x)$ and $f(x)$. }
\label{fig:LwR-overall}
\end{center}
\end{figure}

In this paper, we propose an optimal rejection method for rejecting ambiguous training and test samples for tough character recognition tasks.  
In the proposed method, a rejection function is optimized (i.e., trained) together with a classification function during the training step. 
To realize this co-optimization, we follow the {\em learning-with-rejection} (LwR) framework proposed in theoretical machine learning research~\cite{LwR2016}. Figure~\ref{fig:LwR-overall} illustrates LwR framework. During the training step, several samples are determined to be rejected by a rejection function and they are not used to train the classifier function. It should be emphasized again this is a co-optimization framework where the rejection function and the classification function are optimized by solving a single optimization problem. LwR has many promising characteristics; for example, it has a strong theoretical support about its optimality (about the overfitting risk) and it can use a different feature space for its rejection function.\par
The main contributions of our work are summarized as follows:
\begin{itemize}
    \item This is the first application of LwR to a practical pattern recognition task. Although the idea of LwR is very suitable to various pattern recognition tasks (including, of course, character recognition and document processing), no application research has been done so far, to the authors' best knowledge.
    \item For designing features for classification and rejection, we utilize a convolutional neural network (CNN) as a multiple-feature extractor; the classification function is trained by using the output from the final convolution layer and the rejection function is trained from a different layer.
    \item We conducted two classification experiments to show that LwR outperforms CNN with confidence-based rejection, and traditional SVM with the rejection strategy of Figure~\ref{fig:reject-targets}~(b).
\end{itemize}
\par
The rest of the paper is organized as follows: Section II reviews related works and Section III is devoted to explain the problem formulation and the theoretical background of LwR. In Section IV, the experimental setting and the discussion of final results are shown, while conclusion is given in Section V.
\par

\section{Related work}

Since data usually contains unrecognizable samples, data cleansing  \cite{hernandez1998real,maletic2000data,khayyat2015bigdansing} has been an important topic in many fields such as medical image processing \cite{lehmann1999survey,mcauliffe2001medical}, document analysis \cite{neumann2015efficient,messina2015segmentation,chen2015page} and commercial applications~\cite{rahm2000data}. Among those applications, ambiguous samples in dataset might incur serious troubles at data processing stage. To handle  this problem, many data cleansing methods have been studied \cite{mezzanzanica2015model,lee1999cleansing} and the goal of all of them is to keep the data clean. To guarantee the data integrity,  most of them apply repairing operation to the target data with inconsistent format. Instead of repairing those acceptable flaws, authors in ~\cite{chen2010leveraging} aim to remove erroneous samples from entire data, and in~\cite{fecker2014document}, Fecker et al. propose a method to reject query manuscripts without corresponding writer with a rejection option. These works show the necessity of rejection operation in data cleansing. For machine learning tasks, training data with undesirable samples (as shown in Figure~\ref{fig:reject-targets}~(a)) could lead to a ill-trained model, whose performance in the test phase is doomed to be poor. On the other hand, test data could also contains such unwelcome samples but are apparently better to be removed by the trained model than those data cleansing methods.

Other than data cleansing, rejection operation is also commonly used in many fields of machine learning
such as handwriting recognition \cite{pal2007off,lecun1990handwritten}, medical image segmentation \cite{olabarriaga2001interaction,mekhmoukh2015improved} and image classification tasks \cite{chan2015pcanet,wang2010locality,cirecsan2012multi}. Here, the rejection operation are done during the test stage but not the training stage. In the handwriting recognition field, Kessentini et al. apply a rejection option to Latin and Arabic scripts recognition and gained significant improvements compared to other state-of-the-art techniques ~\cite{kessentini2015dempster}. Mesquita et al. develop a classification method with rejection option~\cite{mesquita2016classification} to postpone the final decision of non-classified modules for software defect prediction. In~\cite{he2009novel}, He et al. construct a model with rejection on handwritten number recognition, where the rejection option nicely prevented misclassifications and showed high reliability. Mekhmoukh and Mokrani employ a segmentation algorithm that is very sensitive to outliers in~\cite{mekhmoukh2015improved} to effectively reject anomalous samples. Niu et al. in~\cite{niu2012novel} present a hybrid model of integrating the synergy of a CNN and SVM to recognize different types of patterns in MNIST, achieved 100\% reliability with a rejection rate of 0.56\%. However, works mentioned above treat rejection operation as the extension of classification function. It is notable that dependence on classification function could endow potential limitation to rejection ability of the models.  

Rejection operation also plays an important role in document processing field  \cite{maitra2015cnn,serdouk2016new}. Bertolami et al. investigate rejection strategies based on various confidence measures in~\cite{bertolami2006rejection} for unconstrained offline handwritten text line recognition. In~\cite{li2016rejecting}, Li et al. propose a new CNN based confidence estimation framework to reject the misrecognized character whose confidence value exceeds a preset threshold. However, to our knowledge, there still has no work on combining CNNs and independent rejection functions in document analysis and recognition field. Since it is well known that CNNs have powerful feature extraction capabilities, outputs of hidden layers in CNNs could be used instead of kernels as feature spaces for model training. In this paper, we propose a novel framework with a rejection function learned along with but not dependent on the classification function, using features extracted from a pre-trained CNN.\par

\newcommand{\bPhi}{\boldsymbol{\Phi}}
\section{Learning with Rejection}

In this section, we introduce the problem formulation of LwR. 
Let $(x_1, y_1)$, $\ldots$ , $(x_m, y_m)$ denote training samples. 
Let $\bPhi$ and $\bPhi'$ denote
functions that map $x$ to different
feature spaces.
Let $f = \boldsymbol{w}\cdot \bPhi(x) +b$ be a 
linear classification function over $\bPhi$,
while $r = \boldsymbol{u} \cdot \bPhi'(x) + b'$ be 
a linear rejection function over $\bPhi'$.
Our LwR problem is formulated as an optimization problem of $\boldsymbol{w}$, $b$, $\boldsymbol{u}$ and $b'$ as follows.
As shown in Figure~\ref{fig:LwR-overall},  a two-class classification problem with reject option uses the following decision rule :
\[
  \begin{cases}
    +1\ (x\ \mathrm{is \ class 1}) & \mathrm{\ if\ } (f(x) >0) \wedge (r(x) >0), \\
    -1\  (x\ \mathrm{is \ class 2}) &\mathrm{\ if\ } (f(x) \leq 0) \wedge (r(x) >0),\\
    \mathrm{reject}  &\mathrm{\ if\ } r(x) \leq 0.
  \end{cases}
\]
\par
To achieve a good balance between 
the classification accuracy and
the rejection rate, we minimize the following risk:
\begin{align}
\label{eq:risk}
    R(f,r)=\sum_{i=1}^m 
    \left(
    1_{(\mathrm{sgn}(f(x_i)) \neq y_i)
    \wedge{(r(x_i)>0)}}
    + c1_{(r(x_i)\leq0)}
    \right),
\end{align}
where $c$ is a parameter
which weights the rejection cost.
The cost represents the penalty to the rejection operation. \par
As an example in document processing task, assuming digit recognition for bank bills.
In this case, there is almost no tolerance to errors and thus the rejection cost $c$ must be set at 
a lower value for rejecting more suspicious patterns to avoid possible mistakes. In contrast, in the case of character recognition of personal diary, $c$ can be set at 
a larger value for accepting more ambiguous characters is acceptable.\par
Since the direct minimization of the risk $R$ is difficult, Cortes et al.~\cite{LwR2016} proposed a convex surrogate loss $L$ as follows:
\[
L=\max \left( 1 + \frac { \alpha } { 2 } ( r ( x ) - y f ( x ) ) , c ( 1 - \beta r ( x ) ) , 0 \right),
\]
where $\alpha=1$ and $\beta=1/(1-2c)$. In~\cite{LwR2016}, it is proved that the minimization problem of $L$ along with  a typical regularization of $\boldsymbol { w }$ and $\boldsymbol { u}$ results in the following SVM-like optimization problem:
\begin{align} 
\label{align:reject_opt_prob}
\nonumber
\min _ { \boldsymbol { w } , \boldsymbol { u },b,b' \boldsymbol { \xi } }\;& \frac { \lambda } { 2 } \| \boldsymbol { w } \| ^ { 2 } + \frac { \lambda ^ { \prime } } { 2 } \| \boldsymbol { u } \| ^ { 2 } + \sum _ { i = 1 } ^ { m } \xi _ { i } \\ \nonumber
\text { sub. to }&\; \xi _ { i } \geq c \left( 1 - \beta \left( \boldsymbol { u } \cdot \boldsymbol { \Phi } ^ { \prime } \left( x _ { i } \right) + b ^ { \prime } \right) \right) , \\ \nonumber
& \; \xi _ { i } \geq 1 + \frac { \alpha } { 2 } \left( \boldsymbol { u } \cdot \boldsymbol { \Phi } ^ { \prime } \left( x _ { i } \right) + b ^ { \prime } - y _ { i }( \boldsymbol { w } \cdot \boldsymbol { \Phi } \left( x _ { i } \right) + b) \right),  \\
& \;\xi _ { i } \geq 0 \; (i=1,\ldots,m),
\end{align}
where $\lambda$ and $\lambda'$ are 
regularization parameters and $\xi_i$ is a slack variable. This is a minimization problem of a quadratic objective function with linear constraints and thus its optimal solution can be easily obtained by using a quadratic programming (QP) solver.\par
It should be emphasized that LwR solution of the above-mentioned formulation has a strong theoretical background. Similarly to the fact that the standard SVM solution has theoretical guarantee of its generalization performance,  the optimal solution of the optimization problem (\ref{align:reject_opt_prob}) in LwR mentioned above
also has a theoretical support. See Appendix for its brief explanation.
It should also be underlined that the original LwR research has mainly focused on this theoretical analysis and thus has never been used for any practical application. To the authors' best knowledge, this paper is the first application of LwR to a practical task.
\begin{figure}[t!]
\begin{center}
\includegraphics[width=\columnwidth]{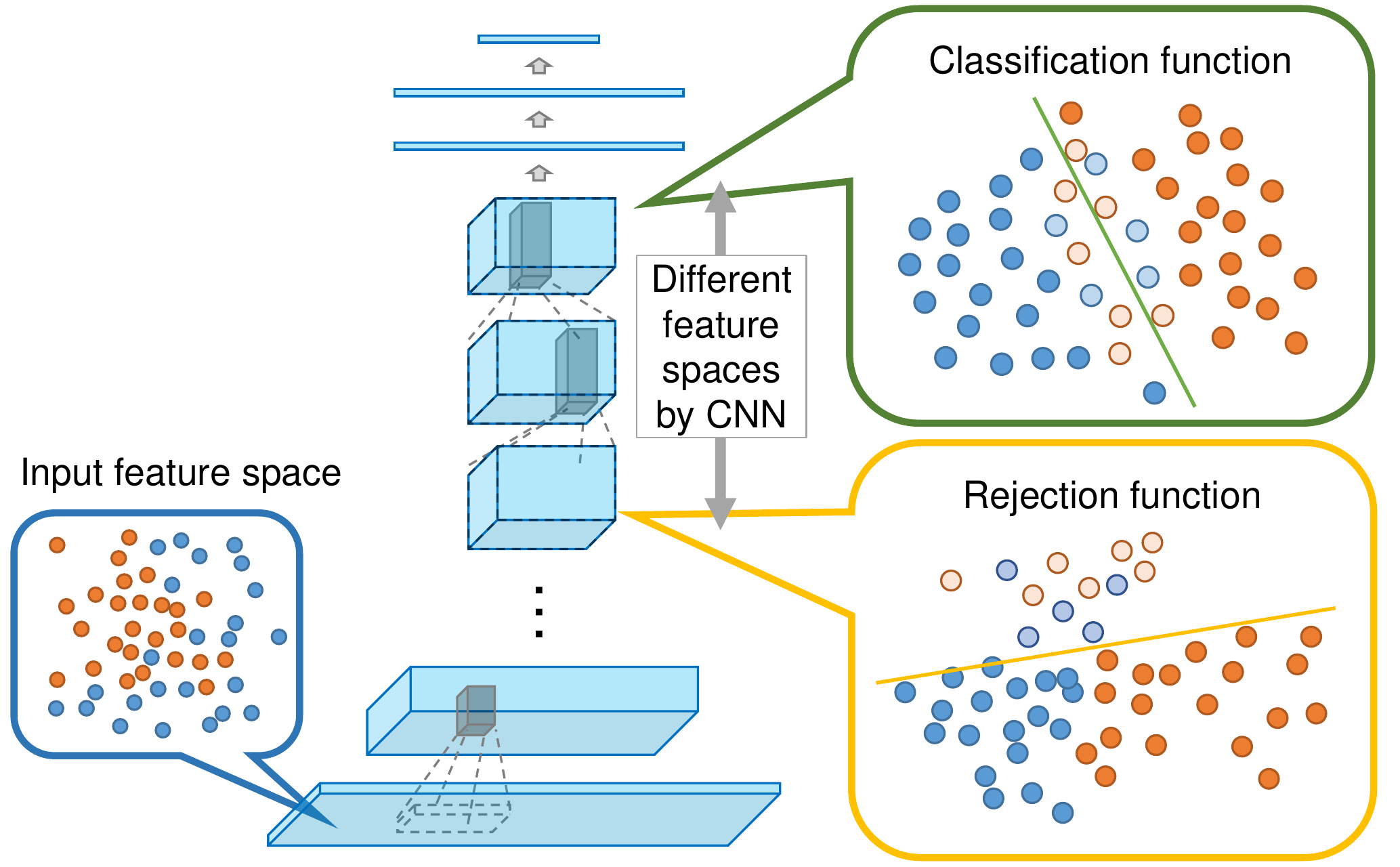}\\[-2mm]
\caption{Extracting different features from CNN for LwR.}
\label{fig:structure}
\end{center}
\end{figure}
\section{Experimental result}
In this section, we demonstrate that our optimal rejection
approach effectively works for character recognition tasks.
More precisely, 
we consider two binary classification tasks: classification of similar characters and classification of character and non-character. 
Of course, in those tasks, it is better to achieve  a high accuracy with a low reject rate.

\subsection{Experimental Setup}
\subsubsection{Methods for comparison}
Through the experiments, we compare three methods: CNN classifier, 
SVM classifier\footnote{We use an implementation of SVM that estimation of posterior class probabilities.}, 
and LwR approach. 
The SVM classifier is set linear for deep CNN features we use already contain nonlinear compositions.

The performances of these three classification models are compared with shifting rejection
cost parameters as $c=0.1,\ldots,0.4$.
CNN and SVM output class probability $p(y|x)$, 
and thus we consider their final output $f(x)$ with reject option 
based on confidence threshold $\theta$, that is,
\[
  f(x)=\begin{cases}
    +1\ (x\ \mathrm{is \ class 1}) & \mathrm{\ if\ } p(+1|x)>\theta,\\
    -1\  (x\ \mathrm{is \ class 2}) &\mathrm{\ if\ } p(-1|x)>  \theta,\\
    \mathrm{reject}  &\mathrm{\ otherwise}.
  \end{cases}
\]
We determine the optimal confidence threshold $\theta$ 
of rejection according to the risk $R$ using the {\em trained} $f(x)$ and a validation set, where the $R$ of CNN and SVM
can be defined as:
\[
    R(f)=\sum_{i=1}^m 
    \left(
    1_{((f(x_i)) \neq y_i) \wedge (f(x)\neq \mathrm{reject})}
    + c1_{(f(x)= \mathrm{reject})}
    \right).
\]
For the LwR model, validation samples are used to tune hyperparameters $\lambda$ and $\lambda'$
for LwR could determine its rejection rule by using only training samples.

\par
The learning scenario
of these three methods in these experiments is introduced as follow.
For each binary classification tasks, we divide the binary labeled training samples into three sets:
(1)~Sample set for training a CNN. This CNN is used as not only the feature extractor for LwR and SVM but also a CNN classifier.
(2)~Sample set for training SVM and LwR with the features extracted by the trained CNN. 
(3)~Sample set for validating the parameters of each method.

\subsubsection{Feature Extraction}
As mentioned above, one of the key ideas of LwR is to use different feature spaces to construct rejection function.
For this point, we use the outputs of two different layers of the trained CNN for the classifier $f(x)$ and rejection function $r(x)$ respectively, as shown in Figure~\ref{fig:structure}. 
Specifically, we use the final convolutional layer of the CNN for classification. Note that the SVM classifier also uses the same layer as its feature extractor.
For the training stage of rejection function, the second final layer of the CNN 
is used for we found it achieved good performance in our preliminary experiments. In addition, if the same layer is used for both classification and rejection functions, the LwR model will degrade to a SVM-like model with information from only one feature space.\par

\subsubsection{Evaluation criteria}
Using sample set $S$ (i.e. the test set) for each task,
we evaluate the classification
performance with reject option
of these three models
by following metrics:
classification accuracy for non-rejected test samples
(which we simply call \emph{accuracy}
for short) and \emph{rejection rate} are defined as:

\begin{figure}[t!]
\begin{center}
\includegraphics[width=\columnwidth]{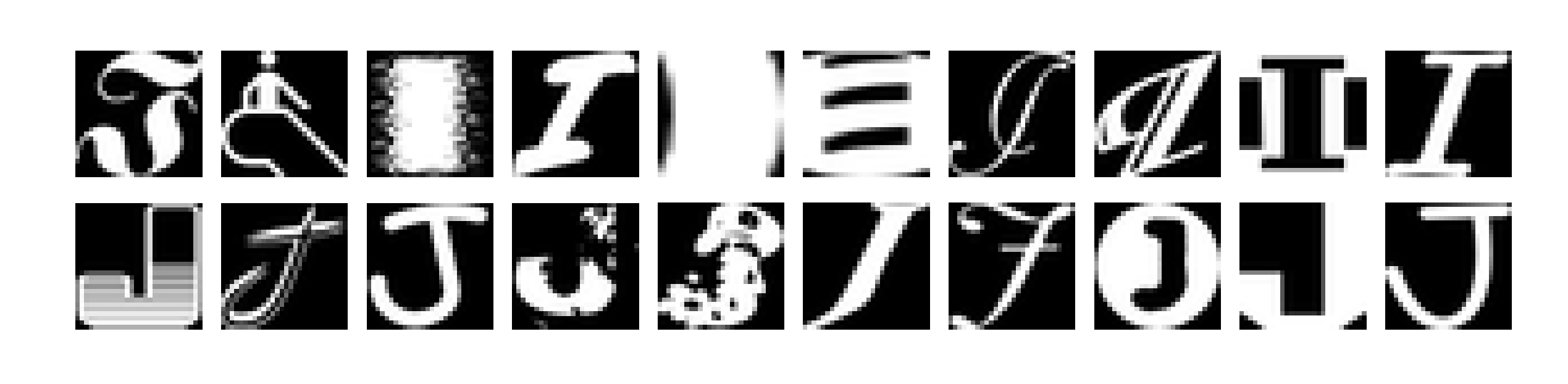}\\[-3mm]
\caption{Examples of ``I'' (upper row) and ``J'' (lower row) form notMNIST.}
\label{fig:notMNIST}
\bigskip
\bigskip
\includegraphics[width=0.5\textwidth]{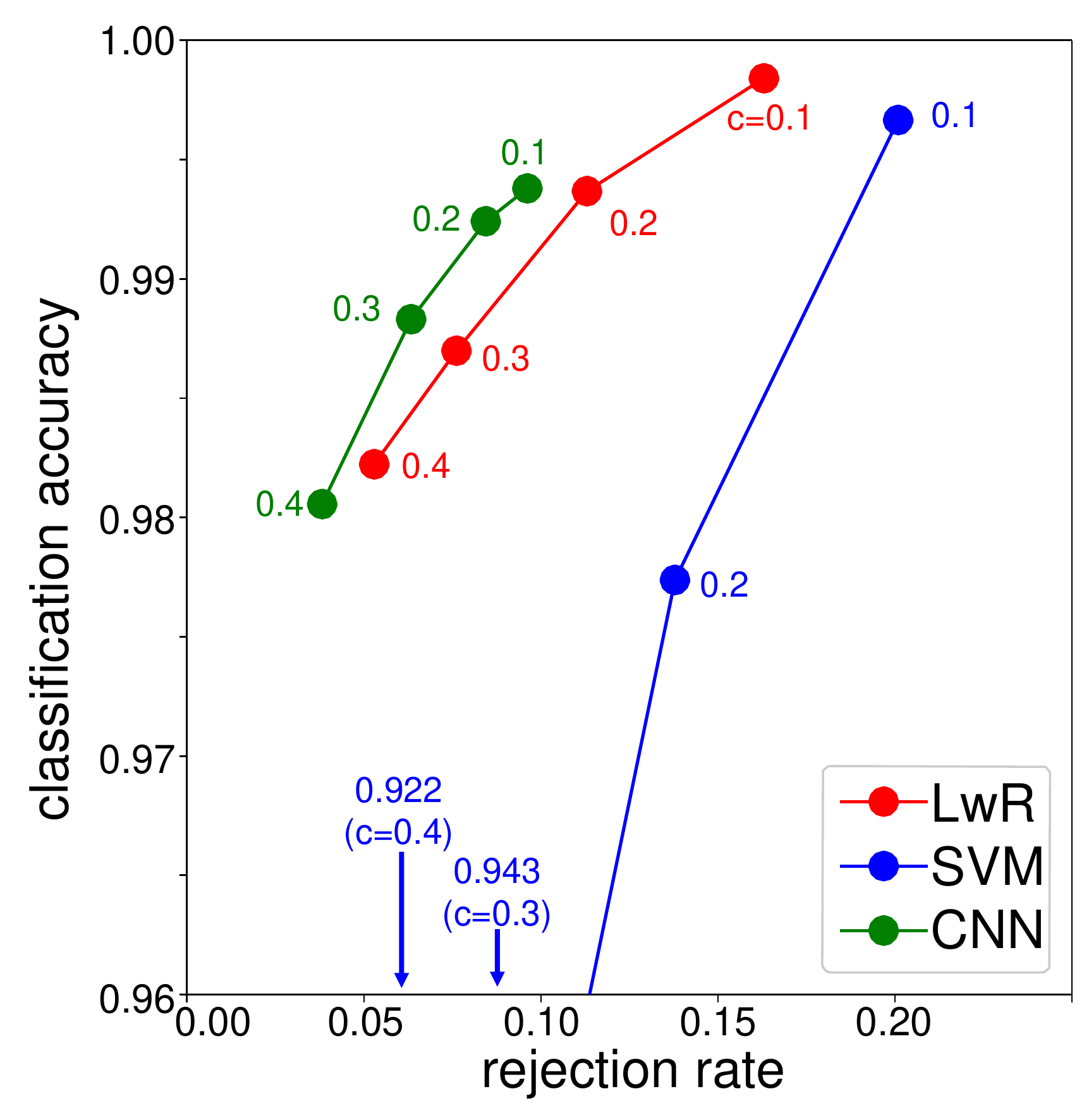}\\[-2mm]
\caption{The relationship between accuracy and reject rate for the classification task between ``I'' and ``J'' from notMNIST dataset.}
\label{fig:notmnist_acc}
\end{center}
\end{figure}

\begin{figure}[t]
\centering
\begin{center}
\centering
\includegraphics[width=\columnwidth]{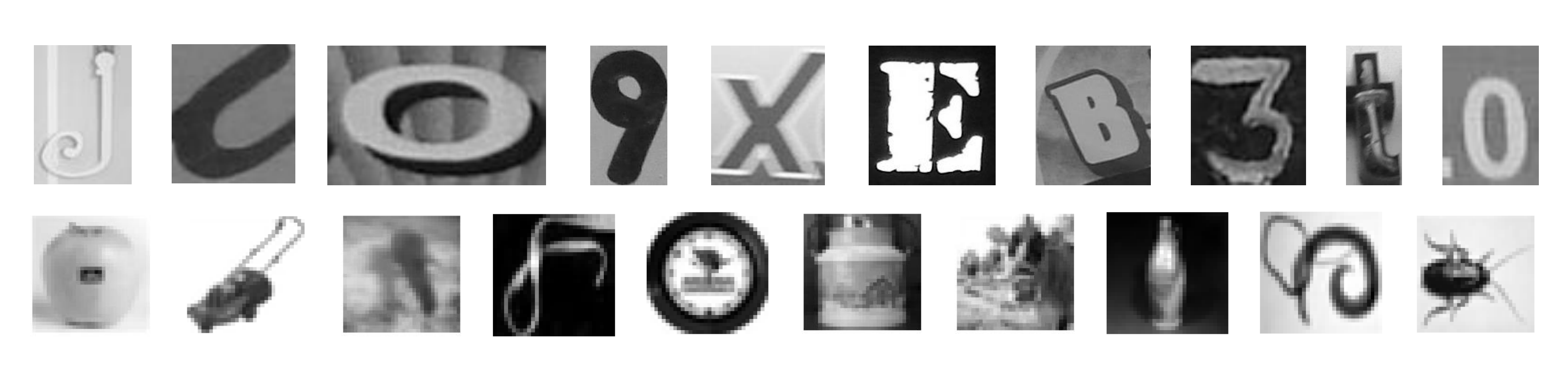}\\[-3mm]
\caption{Examples from Chars74k (upper row) and CIFAR100 (lower row) datasets.}
\label{fig:char74-cifar100}
\bigskip
\bigskip
\includegraphics[width=0.5\textwidth]{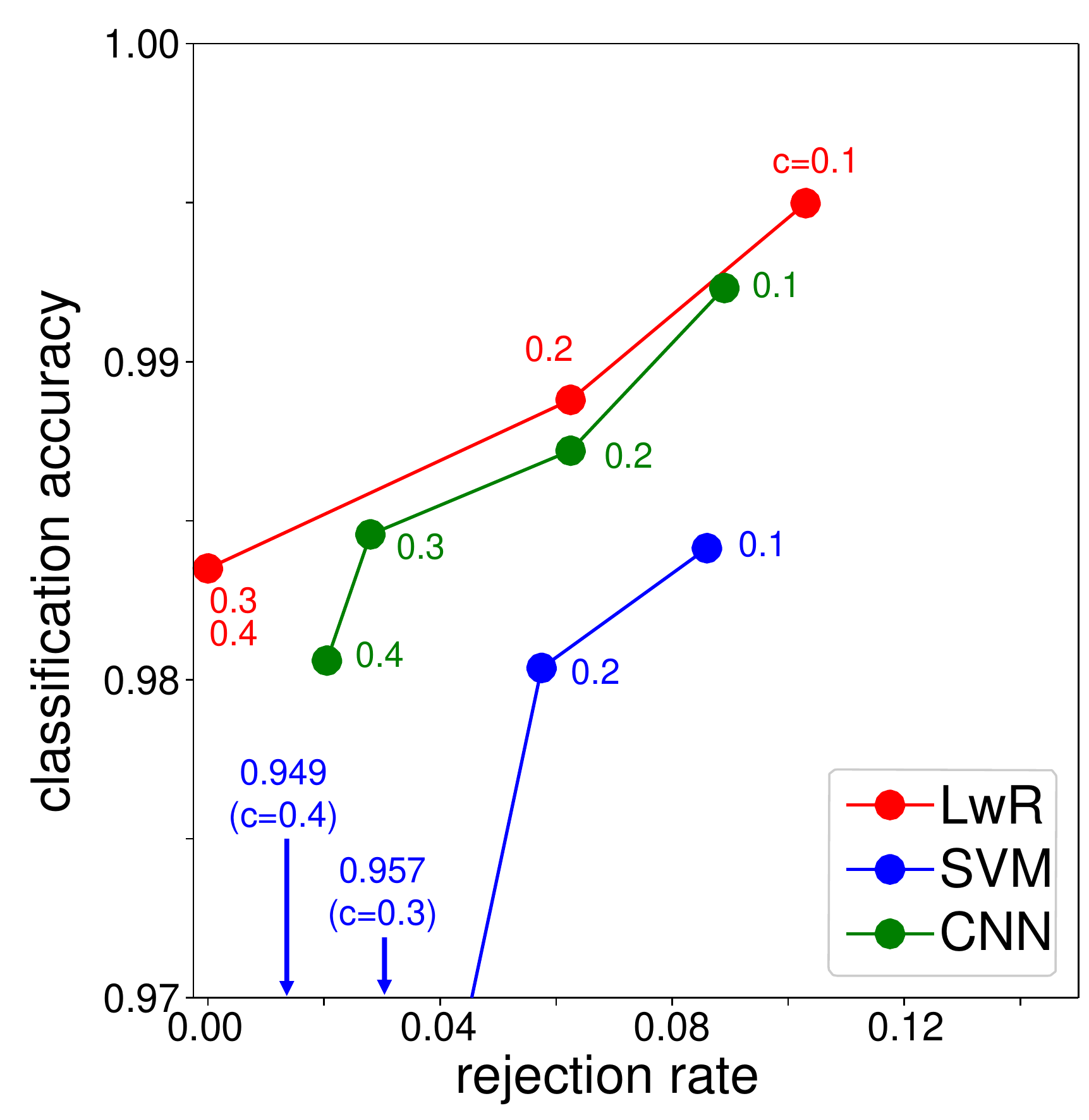}\\[-2mm]
\caption{The relationship between accuracy and reject rate for character and non-character classification using Chars74k and CIFAR100.}
\label{fig:accu}
\end{center}
\end{figure}

\[
\mathrm{accuracy} = \frac{\sum\limits_{x_i\in{\{S-S_{rej}}\}}
1_{f(x_i)=y_i}}{|S| - |S_{\mathrm{rej}}|}, \;\; \mathrm{rejection~rate}= \frac{|S_{\mathrm{rej}}|}{|S|},
\]
where $S_{\mathrm{rej}}$ is the set of rejected samples, 
and $f(x_i)$ is the classification
function of each method.

\subsection{Classification of Similar Characters with Reject Option}
In this section, we use a dataset named notMNIST\footnote{https://www.kaggle.com/lubaroli/notmnist/} as both training and test data.
The notMNIST dataset is an image dataset consists of A-to-J alphabets including various
kinds of fonts, whose examples are shown in Figure~\ref{fig:notMNIST}.
We choose a pair of very similar letters ``I'' and ``J''  to organize a tough two-class recognition task, where reject operation of
many ambiguous samples is necessary.
We randomly selected
3,000 training samples of ``I'' and ``J'' respectively from notMNIST dataset.
We also used 2,000 samples for CNN training, 2,000 samples for SVM and LwR training, and 2,000 samples for validations. We then obtain CNN classifier and SVM with reject option, and LwR. The CNN architecture in this experiment is as follows:
two convolutional layers (3 kernels, kernel size $3 \times 3$ with stride 1), two pooling layers (kernel size $2 \times 2$ with stride 2) with Rectified Linear Units (ReLU) as the activation function, three fully-connected layers (3,136, 1,568, and 784 hidden units respectively), with minimization
of cross-entropy loss. 

Figure~\ref{fig:notmnist_acc} shows the relationship between accuracy and reject rate, evaluated by using the test dataset. LwR clearly shows significantly better performance than the standard SVM with rejection option. Surprisingly, CNN could show slightly better performance than LwR at the stage when the rejection rate is low (although LwR showed the highest accuracy in this experimental range). This might be because the classifier of CNN, i.e., fully-connected layers of CNN is trained (or tuned) together with the feature extraction part of CNN and thus CNN has an advantage by itself. We will see that LwR outperforms in a more realistic experiment shown in the next section.

\begin{figure}
\begin{center}
\includegraphics[width=\columnwidth]{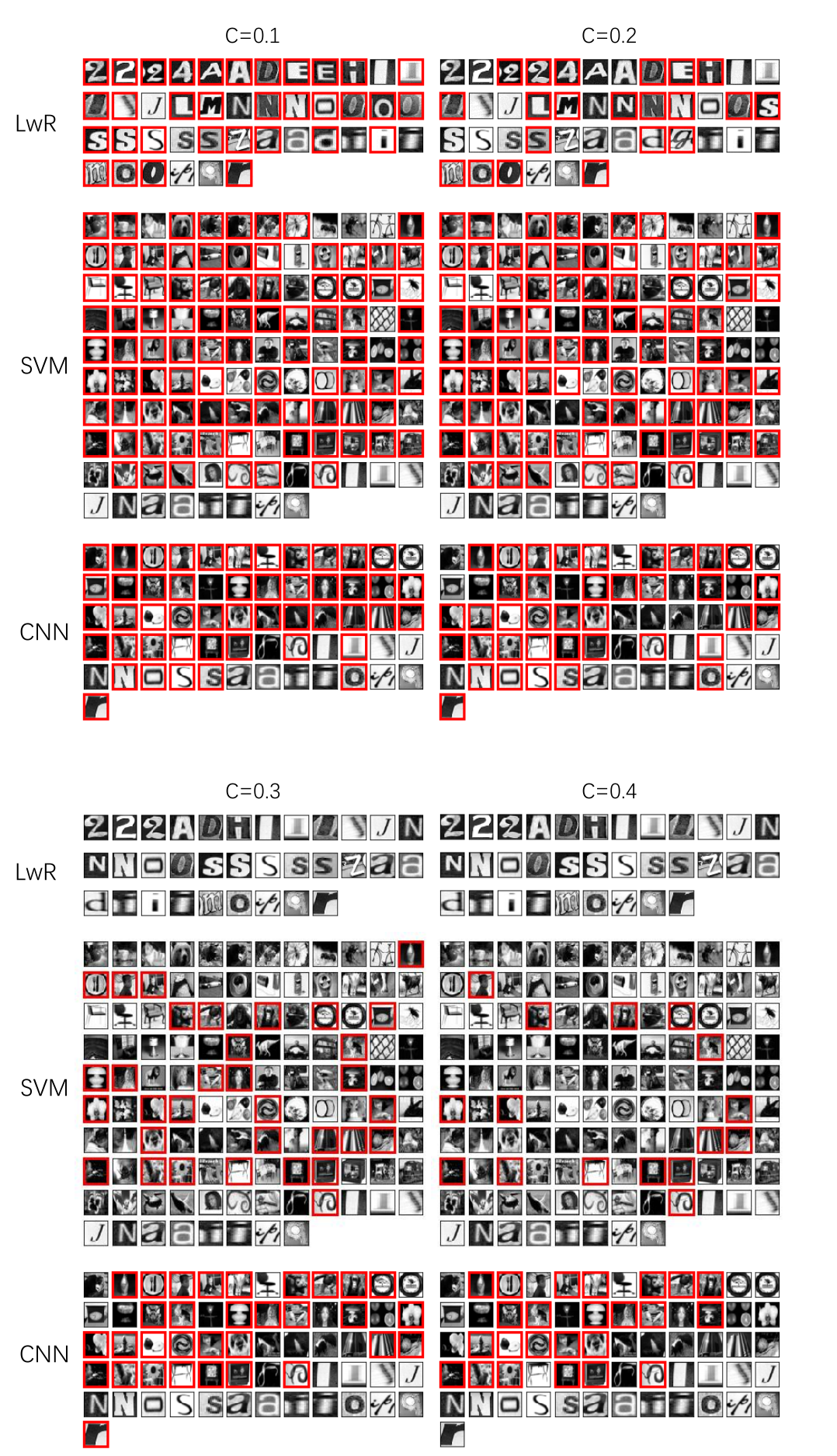}\\[-5mm]
\caption{Test samples misclassified by the LwR, the standard SVM and the CNN at different $c\in\{0.1, 0.2 ,0.3 ,0.4\}$. Rejected samples are highlighted in red. }
\label{fig:wrong samples}
\end{center}
\end{figure}

\subsection{Classification of Character and Non-character with Reject Option}
As a more realistic application, we consider the classification of character and non-character.
It is known that this classification task is a part of scene text detection task and still a tough recognition task (e.g.,~\cite{bai2017text}) because of the ambiguity between  characters and non-characters.
In this experiment, 
as a scene character image dataset,
we use Chars74k dataset\footnote{http://www.ee.surrey.ac.uk/CVSSP/demos/chars74k/}, which contains large number of character images in the natural scene.
As a non-character image dataset,
we use CIFAR100 datset\footnote{https://www.cs.toronto.edu/~kriz/cifar.html}
which contains various
kinds of objects in the
natural scene. The image samples from those datasets were converted into gray scale and resized into $32 \times 32$ pixels.
Figure~\ref{fig:char74-cifar100} shows several examples. \par

We randomly sample 5,000 images respectively from both CIFAR100 and Chars74k datasets. Among them, 6,000 are used for CNN training, 2,000 for SVM and LwR training, and 2,000 samples for validations. After training,
we obtain CNN classifier and SVM with rejection thresholds,
and LwR. CNN with the following architecture was used for this task:
convolutional layers (32, 64, 128 kernels, kernel size $3 \times 3$, $5 \times 5$, $5 \times 5$, with stride 1, 2, 2 respectively, using ReLU as activation function), one max pooling layer ($2 \times 2$ sized kernel with stride 2) and two fully-connected layers (512 and 2 units with ReLU and softmax), with minimization of cross entropy.
\par
Figure~\ref{fig:accu} shows the test accuracy using 2,000 samples
randomly chosen from Chars74k and CIFAR100\footnote{CIFAR100 has much more samples for testing, but unfortunately, Chars74k remains highly limited number of samples for testing.}.
We can see LwR could achieve the best compromise between accuracy and rejection rate among three methods.
Surprisingly, when $c=0.3$ and
$c=0.4$, LwR predicts the labels of all test samples
(i.e., no rejected test samples) with keeping 
a higher accuracy than CNN.
This result confirm the strength of theoretically formulated LwR
that obtains the classification 
and rejection function for minimizing the risk $R$,
while the rejection option for CNN and SVM is rather heuristic. \par
%


Figure~\ref{fig:wrong samples} shows the image samples are
to be misrecognized without the rejection function. Namely, 
if we only rely on $f(x)$, the sample $x$ is misrecognized.
Among them, samples highlighted by red boxes are samples {\em successfully rejected} by the rejection function $r(x)$.
Those highlighted samples represent that classification function $f(x)$ and rejection function $r(x)$ work in a complementary manner. 
In addition, we can find
LwR has a lot of different rejected samples from SVM and CNN,
whereas SVM and CNN share a lot of common samples.
This result is induced by the effect that LwR employs different feature spaces for classification and rejection.

\section{Conclusion}
We propose an optimal rejection method for character recognition tasks. The rejection function is 
trained together with a classification function under the framework of Learning-with-Rejection (LwR). One technical highlight is that the rejection function can be trained on an arbitrary feature space which is different from the feature space for the classification. Another highlight is that LwR is not heuristic and its performance is guaranteed by a machine learning theory. From the application side, this is the first trial of using LwR for practical tasks. The experimental results 
show that the the optimal rejection is useful for tough character recognition tasks 
than traditional strategies, such as SVM with a threshold. Since LwR is a general framework, we believe that
it will be applicable to more tough recognition tasks and possible to achieve stable performance by its rejection function. \par
As future work,
we plan to extend our framework to
more realistic problem setting
such as multi-class setting
by using~\cite{NIPS1999_1773}.
Furthermore, we will consider
co-training framework
of classification feature space and rejection feature space,
which enables the truly optimal classification
with reject option.
\par
\par
\section*{Acknowledgement}
This work was supported by JSPS KAKENHI Grant Number JP17H06100 and JP18K18001.
\section*{Appendix}
The optimal solution of (\ref{align:reject_opt_prob})
has a theoretical guarantee of the generalization performance as follows:
\begin{theorem}[proposed by~\cite{LwR2016}]
Suppose that we choose classification function $f$ from 
a function set $F$ and choose rejection function $r$ 
from a function set $G$. We denote $R$ for test samples by $R_{test}$,
and $R$ for training samples by $R_{train}$.
Then, the following holds:
\begin{align}
    R_{test}(f,r) \leq R_{train}(f, r) + \mathfrak{R}(F) + (1+c)\mathfrak{R}(G).
\end{align}
\end{theorem}
In above, $\mathfrak{R}$ is the Rademacher complexity~\cite{Bartlett:2003:RGC},
which is a measure evaluating the theoretical overfitting risk of a function set. 
Roughly speaking, this theorem says that, if we prepare a proper feature space
(i.e., not-so-complex feature space) mapped by $\bPhi$ and $\bPhi'$, 
we will achieve a performance on test samples as high as on training samples.
Thus, by solving the problem~(\ref{align:reject_opt_prob})
of minimizing $R_{train}$,
we can obtain theoretically optimal classification and
rejection function which achieves the best balance
between the classification error for not-rejected samples
and the rejection cost~\cite{maitra2015cnn}.

\newcommand{\BIBdecl}{\setlength{\itemsep}{0.2mm}}
\bibliographystyle{IEEEtran}
\bibliography{samplepaper}




\end{document}